\documentclass[10pt,twocolumn,letterpaper]{article}

\usepackage[pagenumbers]{cvpr} 

\usepackage[accsupp]{axessibility}  

\usepackage{multirow}
\usepackage{graphicx}
\usepackage{colortbl}
\usepackage{float}
\usepackage[symbol]{footmisc}
\usepackage{afterpage}
\usepackage[dvipsnames]{xcolor}
\definecolor{mygray}{gray}{0.6}

\definecolor{cvprblue}{rgb}{0.21,0.49,0.74}
\usepackage[pagebackref,breaklinks,colorlinks,allcolors=cvprblue]{hyperref}

\def\paperID{8} 
\def\confName{CVPR}
\def\confYear{2025}

\title{DCSEG: Decoupled 3D Open-Set Segmentation using Gaussian Splatting}

\author{Luis Wiedmann$^*$ \hfill Luca Wiehe$^*$ \hfill David Rozenberszki \\
Technical University of Munich\\
{\tt\small \{luis.wiedmann, luca.wiehe, david.rozenberszki\}@tum.de}}

\begin{document}
\twocolumn[{%
\renewcommand\twocolumn[1][]{#1}%
\maketitle
\begin{center}
    \centering
    \captionsetup{type=figure}
    \includegraphics[width=\linewidth]{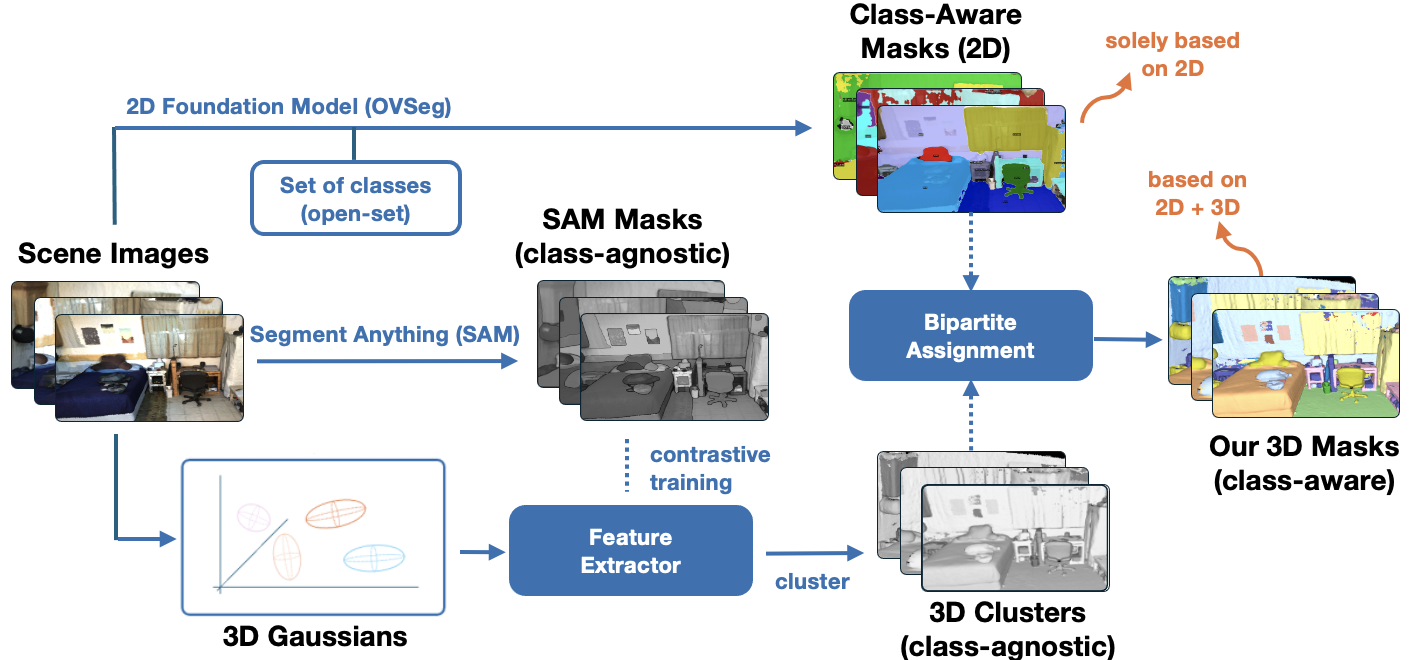}
    \captionof{figure}{\textbf{Decoupling the semantic segmentation pipeline.} 
    We present DCSEG, a holistic 3D reconstruction and scene understanding method. At the core of our method, we leverage pre-trained 2D foundation models to recognize uniform semantic concepts in 2D images of 3D scenes and use these predicted masks as contrastive optimization targets from multi-view images to class-agnostic 3D instances and object parts. These features are then used to cluster the Gaussians in 3D with hierarchical clustering methods. Simultaneously, we use a 2D semantic segmentation network to obtain class-aware masks and aggregate class-agnostic parts into meaningful semantic instances. As a result, we obtain 2D/3D instance and semantic segmentation on synthetic and real-world scenes. 
    }
    \label{fig:our_method}
\end{center}%
}]

\begin{abstract}
Open-set 3D segmentation represents a major point of interest for multiple downstream robotics and augmented/virtual reality applications. 
We present a decoupled 3D segmentation pipeline to ensure modularity and adaptability to novel 3D representations as well as semantic segmentation foundation models. We first reconstruct a scene with 3D Gaussians and learn class-agnostic features through contrastive supervision from a 2D instance proposal network. These 3D features are then clustered to form coarse object- or part-level masks. Finally, we match each 3D cluster to class-aware masks predicted by a 2D open-vocabulary segmentation model, assigning semantic labels without retraining the 3D representation. Our decoupled design (1) provides a plug-and-play interface for swapping different 2D or 3D modules, (2) ensures multi-object instance segmentation at no extra cost, and (3) leverages rich 3D geometry for robust scene understanding. We evaluate on synthetic and real-world indoor datasets, demonstrating improved performance over comparable NeRF-based pipelines on mIoU and mAcc, particularly for challenging or long-tail classes. We also show how varying the 2D backbone affects the final segmentation, highlighting the modularity of our framework. These results confirm that decoupling 3D mask proposal and semantic classification can deliver flexible, efficient, and open-vocabulary 3D segmentation.
\end{abstract}
\footnotetext{* Equal contribution. Code available \href{https://github.com/lusxvr/dcseg}{here}.}
\section{Introduction}
Understanding the semantic and instance-level structure of 3D scenes is a key requirement in various downstream applications, including robotics, augmented/virtual reality, and autonomous driving.
Recent progress in Neural Radiance Fields (NeRFs) \cite{mildenhall2021nerf} has enabled impressive quality in novel-view synthesis and 3D scene capture. However, NeRF-based approaches typically require volumetric rendering, which is computationally expensive and can be less flexible for certain real-time applications. In contrast, 3D Gaussian Splatting (3DGS) \cite{kerbl20233d}, and its follow-ups, offer an explicit representation of the scene through a set of 3D Gaussian primitives. By rasterizing these Gaussians directly onto the image plane, we can achieve much faster rendering.

Despite the development of these new representations, the problem of open-vocabulary 3D semantic segmentation remains challenging. Unlike closed-set 3D segmentation methods that assume a fixed set of classes, open-vocabulary methods aim to handle broad or arbitrary category labels, often by leveraging large‐scale vision–language pretraining. This is especially beneficial in environments where unexpected or tail classes appear.
In 2D, methods such as CLIP \cite{radford2021clip}, OpenSeg \cite{ghiasi2022scaling}, and OVSeg \cite{liang2023open} map pixels into semantically rich feature spaces that can be queried by textual prompts. Techniques like LERF \cite{kerr2023lerf} transfer these open-vocabulary features into a 3D NeRF representation, while OpenScene \cite{peng2023openscene} combines language embeddings with 3D feature fusion from multi-view data. SAGA \cite{cen2023saga} builds on Gaussian Splatting and lifts 2D features to 3D space via a contrastive optimization, to enable semantic clustering of the underlying Gaussians.
A key challenge for both closed- and open-vocabulary 3D segmentation pipelines is how to robustly incorporate rich geometry with generalizable semantic priors, often learned from large 2D image datasets. Conventional 3D networks (e.g., MinkowskiNet \cite{choy20194d}) require labeled 3D data, which is scarce and expensive to collect.  
Other approaches \cite{peng2023openscene, kerr2023lerf}, fuse 3D structure with language‐conditioned 2D embeddings, enabling semantic queries in an open‐vocabulary manner.
However, these methods are often coupled to the underlying 3D representation (e.g., NeRFs) or rely on point clouds with sparse geometry, restricting their flexibility.

In this paper, we present DCSEG, a decoupled 3D open-vocabulary segmentation pipeline designed around 3D Gaussian Splatting. The key insight is to separate the mask proposal (class-agnostic clustering in 3D) from the mask classification (assigning class labels via 2D foundation models). Concretely, we first learn compact 3D features for each Gaussian using contrastive learning signals from a 2D instance proposal model (e.g., SAM \cite{yang2023sam3d}) and then cluster these features into instance-level or part-level segments in 3D. Next, to achieve open-vocabulary labeling, we match these 3D clusters to class-aware masks derived from large-scale 2D segmentation backbones such as OVSeg \cite{liang2023open} or OpenSeg \cite{ghiasi2022scaling}.
We evaluate our approach on both synthetic (Replica \cite{straub2019replica}) and real-world (ScanNet \cite{dai2017scannet}) datasets. Our results show competitive performance, especially in how the proposed method can segment instances in 3D with minimal confusion in large or repetitive surfaces. Additionally, our method generates insights into the instance- or part-level structure of the scene without specialized training or adaptation.
Our contributions can be summarized as follows:
\begin{itemize}
    \item We utilize 3D Gaussian Splatting as an underlying representation for class-aware open-vocabulary semantic scene segmentation
    \item We demonstrate that Gaussian Splatting can outperform comparable SOTA NeRF-based architectures for 3D semantic segmentation while being more modular
    \item We present an architecture that can identify 3D instances and event parts without needing to train an instance-segmentation network
\end{itemize}
\section{Related Work}

\paragraph{3D Semantic and Instance Segmentation.}
Classical point cloud networks (e.g., MinkowskiNet \cite{choy20194d}) or voxel-based approaches (e.g. VoxelNet \cite{zhou2018voxelnet}) for semantic segmentation rely on fully-supervised training with 3D-labeled data, such as those from large datasets like ScanNet \cite{dai2017scannet}. More recently, NeRF-based segmentation methods, including Panoptic-NeRF \cite{fu2022panoptic} and OpenNeRF \cite{engelmann2024opennerf}, exploit the volumetric rendering pipeline to fuse semantic cues with novel-view generation. A key challenge for the application of volumetric rendering-based pipelines in real-world scenarios is the absence of explicit geometry. One example is navigation in robotics, where the explicit geometry can be used to efficiently perform obstacle avoidance \cite{lei2024gaussnav, chen2024splatnav}.
Alongside semantic segmentation, approaches like Segment3D \cite{huang2024segment3d} or UnScene3D \cite{rozenberszki2024unscene3d} leverage unsupervised or weakly supervised signals to segment instances in 3D. Meanwhile, SAI3D \cite{yin2024sai3d} and OpenMask3D \cite{takmaz2023openmask3d} propose class-agnostic 3D masks, then assign labels a posteriori. The majority of these methods operates on point clouds or voxel grids. These representations become impractical as scene complexity grows, with point clouds requiring dense sampling to capture details, leading to memory bottlenecks, and voxel grids facing cubic storage and computation costs. This trade-off limits their use in high-resolution or large-scale scenes.
3DGS explicitly represents scenes as 3D Gaussians, enabling direct access to geometric structure for tasks like segmentation and collision avoidance. Additionally, its splatting-based rendering is more efficient than voxel grids, allowing high-resolution processing without sacrificing detail.

\paragraph{3DGS-based Semantic Segmentation.}
Recent work explores semantic segmentation within 3DGS frameworks. Semantic Gaussians \cite{guo2024semantic} projects CLIP \cite{radford2021clip} features into 3D space or integrates Gaussian parameters into point-cloud segmentation backbones, but inherits noise from 2D feature lifting. Langsplat \cite{qin2024langsplat} distills multi-resolution SAM masks with CLIP embeddings into a compressed latent space tied to 3D Gaussians, while Feature 3DGS \cite{zhou2024feature3dgs} employs student-teacher distillation from 2D foundation models. However, these methods are tightly coupled to specific embedding spaces (e.g., CLIP) or foundation models, requiring retraining when switching models. 
In contrast, our approach decouples 3D clustering from 2D feature extraction, enabling modular integration of any vision-language model (e.g., OpenSeg \cite{ghiasi2022scaling}, OVSeg \cite{liang2023open}) at inference without retraining. By first establishing a geometrically consistent, language-independent class-agnostic segmentation of 3D Gaussians, we provide a robust foundation for subsequent labeling—this avoids propagating language model ambiguities into the segmentation itself while enabling compatibility with any language model for post-hoc mask classification and differentiates us from existing approaches.

\begin{figure*}[h]
\begin{center}
\begin{tabular}{c c c c}
    \includegraphics[width=0.225\textwidth]{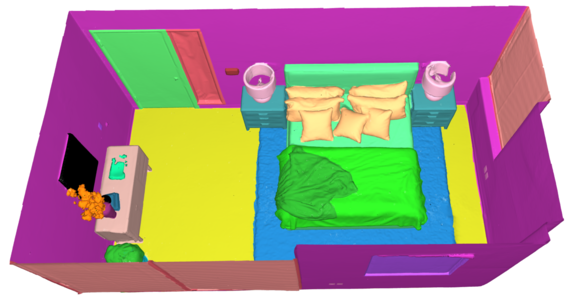} & 
    \includegraphics[width=0.225\textwidth]{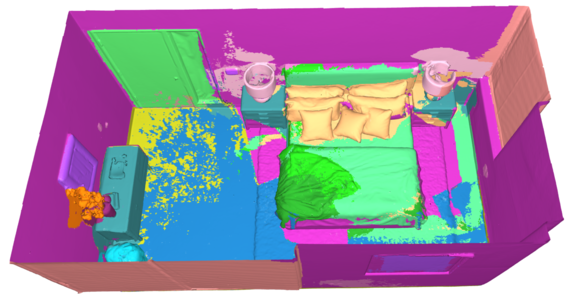} & 
    \includegraphics[width=0.225\textwidth]{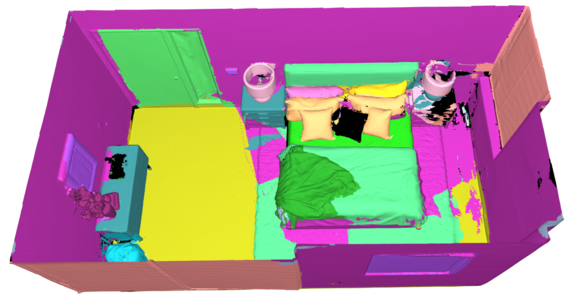} & 
    \includegraphics[width=0.225\textwidth]{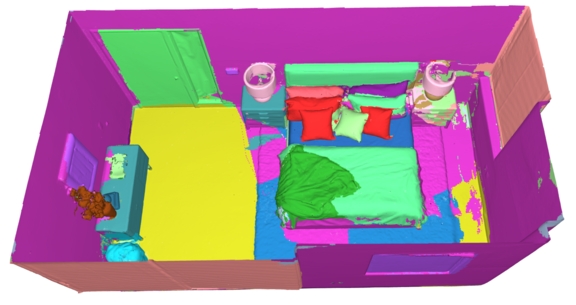} \\
    \includegraphics[width=0.225\textwidth]{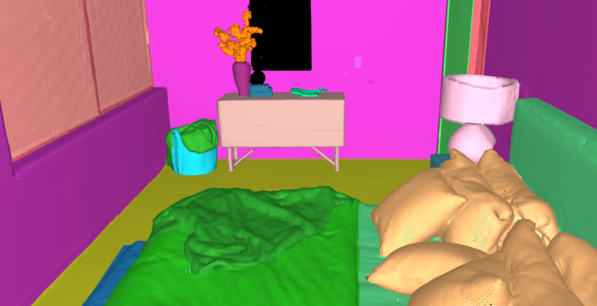} & 
    \includegraphics[width=0.225\textwidth]{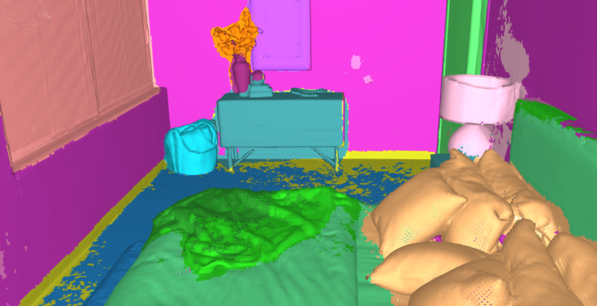} & 
    \includegraphics[width=0.225\textwidth]{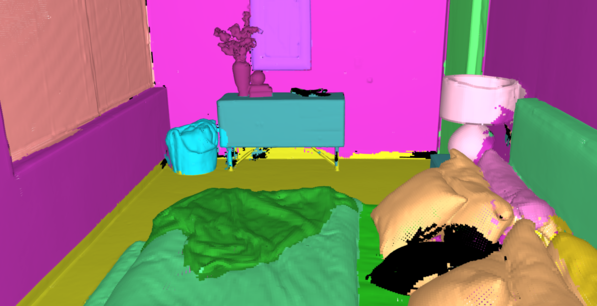} & 
    \includegraphics[width=0.225\textwidth]{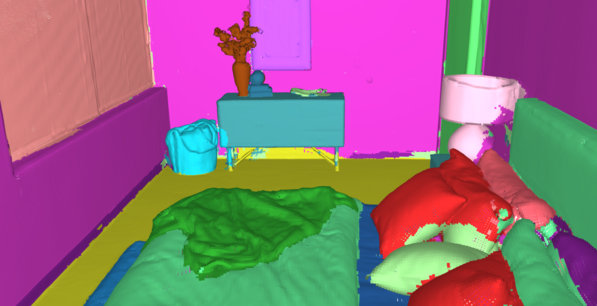} \\
    Ground Truth & OpenNeRF & DCSEG (Openseg) & DCSEG (OVSeg)
\end{tabular}
\end{center}
\caption{\textbf{Segmentation results of our method (DCSEG) compared to the ground truth and OpenNeRF.} Our segmentation masks can detect boundaries more accurately e.g. the blanket/pillows or the wall behind the bed-lamps. Large uniform areas, such as the floor, can be detected with significantly less noise. Switching between Openseg and OVSeg can be done without retraining and demonstrates adaptability with respect to foundation models.}
\label{fig:results_img}
\end{figure*}

\paragraph{Decoupled Segmentation.} Decoupled segmentation approaches separate mask proposal from mask classification, enabling independent optimization of each stage. While 2D methods like DeOP \cite{han2023deop} and ZegFormer \cite{ding2022zegformer} demonstrate the benefits of such an architecture, relying solely on 2D images loses essential 3D contextual information present in real-world scenarios. OpenMask3D \cite{takmaz2023openmask3d} and SAI3D \cite{yin2024sai3d} use point-based representations to lift this paradigm to 3D. Given the aforementioned limitations of point cloud representations, our 3DGS-based alternative will also be an improvement to existing methods in the decoupled segmentation domain.

\begin{table*}
\centering
\resizebox{0.9\textwidth}{!}{%
\begin{tabular}{lcccccccc}
\toprule
 &
  \multicolumn{2}{c}{\textit{Total}} &
  \multicolumn{2}{c}{\textit{Head}} &
  \multicolumn{2}{c}{\textit{Common}} &
  \multicolumn{2}{c}{\textit{Tail}} \\ 
  \cmidrule(r){2-3} \cmidrule(r){4-5} \cmidrule(r){6-7} \cmidrule(r){8-9}
 &
  \cellcolor[HTML]{EFEFEF}mIoU & mAcc &
  \cellcolor[HTML]{EFEFEF}mIoU & mAcc &
  \cellcolor[HTML]{EFEFEF}mIoU & mAcc &
  \cellcolor[HTML]{EFEFEF}mIoU & mAcc \\ \midrule
LERF \cite{kerr2023lerf} &
  \cellcolor[HTML]{EFEFEF}10.5 & 25.8 &
  \cellcolor[HTML]{EFEFEF}19.2 & 28.1 &
  \cellcolor[HTML]{EFEFEF}10.1 & 31.2 &
  \cellcolor[HTML]{EFEFEF}2.3 & 17.6 \\
OpenScene \cite{peng2023openscene} &
  \cellcolor[HTML]{EFEFEF}15.9 & 24.6 &
  \cellcolor[HTML]{EFEFEF}31.7 & 44.8 &
  \cellcolor[HTML]{EFEFEF}14.5 & 22.6 &
  \cellcolor[HTML]{EFEFEF}1.5 & 6.3 \\
OpenNeRF \cite{engelmann2024opennerf} &
  \cellcolor[HTML]{EFEFEF}19.1 & 32.1 &
  \cellcolor[HTML]{EFEFEF}30.5 & 44.2 &
  \cellcolor[HTML]{EFEFEF}\textbf{20.2} & 33.5 &
  \cellcolor[HTML]{EFEFEF}6.6 & 18.6 \\ \midrule
DCSEG (Ours) &
  \cellcolor[HTML]{EFEFEF}\textbf{19.9} & \textbf{33.1} &
  \cellcolor[HTML]{EFEFEF}\textbf{38.1} & \textbf{47.6} &
  \cellcolor[HTML]{EFEFEF}16.1 & \textbf{34.4} &
  \cellcolor[HTML]{EFEFEF}\textbf{6.7} & \textbf{19.3} \\ \bottomrule
\end{tabular}%
}
\caption{\textbf{3D Semantic Segmentation scores on Replica \cite{straub2019replica} with reproducible results from LERF, OpenScene, and OpenNeRF.} The \textit{Total} is over all 51 classes, with the \textit{Head}, \textit{Common}, and \textit{Tail} splits defined following OpenNeRF, each consisting of one-third of the total labels with 17 classes each.}
\label{tab:segmentation_results}
\end{table*}
\section{Method}
Each of the approaches above faces at least one of the following weaknesses: the inability to perform class-aware segmentation, the inability to incorporate (dense) 3D information, the inability to distinguish instances or the coupling between semantic segmentation and 3D reconstruction.
We aim to compensate for all these weaknesses and develop a robust and modular approach to perform 3D open-set segmentation in a class-aware fashion. We seek to achieve this through a decoupled approach, allowing us to interchange the underlying 3D Representation and the semantic feature extraction with any other pipeline that can provide class-agnostic 3D clustering and class-aware 2D segmentation. Our pipeline consists of two essential stages:
\begin{enumerate}
    \item Propose class-agnostic segmentation masks that are based on a 3D representation.
    \item Classify these class-agnostic masks by establishing correspondence with multiple-view class-aware 2D segmentation masks.
\end{enumerate}

\paragraph{Stage 1: Class-Agnostic Mask Proposal.}
Given a set $\mathcal{I}$ of posed RGB-D input images of a 3D scene, we start by obtaining a 
3D reconstruction using Gaussian Splatting. This results in a set of $k$ Gaussians $\mathcal{G} = \{\mathbf{g}_i\}_{i=1..k}$ 
representing the scene.
Inspired by SAGA \cite{cen2023saga}, we then use a scale-aware contrastive learning strategy to attach a set of Gaussian 
affinity features $\mathcal{F} = \{\mathbf{f_{g_i}} \mid \mathbf{f_{g_i}} \in \mathbb{R}^n \}_{i=1..k}$ 
to every Gaussian. Let $\mathbf{p_1}$ and $\mathbf{p_2}$ be two corresponding pixels from a given image
$\mathbf{I} \in \mathcal{I}$, then the loss function is given by: $$\mathcal{L} = \sum_{\mathbf{p_1}, 
\mathbf{p_2}} \mathcal{L}_{corr}(s, \mathbf{p_1}, \mathbf{p_2}) + \frac{1}{h \cdot w} \sum_{\mathbf{p}} 
\mathcal{L}_{norm}(\mathbf{p})$$
This loss contains two main components: A correspondence distillation loss $\mathcal{L}_{corr}$ and a feature normalization loss $\mathcal{L}_{norm}$. The correspondence distillation loss resembles the optimization target that two pixels $\mathbf{p_1}, \mathbf{p_2}$ from a given image $\mathbf{I} \in \mathcal{I}$ should have similar features if and only if they belong to the same SAM mask. 
Note that these features are conditioned on a scale hyperparameter $s$. This hyperparameter is geared towards preserving SAM's granularity. This allows us to adjust the level of detail that is supposed to be captured without the need to rerun the feature extraction.
The normalization loss aims to prevent misalignment between the 2D projected features and the 
original 3D features. It achieves this by imposing a constraint on the norm of the feature vector.
For further details regarding the loss formulation and refinement, refer to \cite{cen2023saga}.

Once each Gaussian $\mathbf{g_i} \in \mathcal{G}$ has a corresponding feature $\mathbf{f_{g_i}}$ 
attached to it, we can use these features 
for clustering. We apply a density-based hierarchical clustering algorithm (HDBScan) \cite{Campello2013DensityBasedCB} that can be formally described as a function $f(\mathbf{f_{g_i}}) \rightarrow \{ 1, 2, ..., M \}$ where $M$ describes the total number of clusters identified by HDBScan. 
In anticipation of the mask classification stage, we rasterize these clusters back onto 2D to obtain binary 2D segmentation masks. These frames are rasterized from the same perspective as the set of input images $\mathcal{I}$. As a result, we obtain the set of masks $\mathcal{M}_a \in \{0,1\}^{M \times h \times w}$ for every input image $\mathbf{I} \in \mathcal{I}$, consisting of $M$ class-agnostic binary masks. 

\paragraph{Stage 2: Mask Classification.}
Once we have obtained the set of projected 3D-based class-agnostic masks \(\mathcal{M}_a\), we need to assign a semantic class label to each of the 3D clusters. We do this using a simple yet effective assignment method. We utilize a 2D foundation model (e.g. OVSeg \cite{liang2023open} or OpenSeg \cite{ghiasi2022scaling}) for mask classification of the $N$ classes in a given input image $\mathbf{I} \in \mathcal{I}$, generating a set of class-aware masks $\mathcal{M}_b \in \{0,1\}^{N \times h \times w}$ in 2D space. Each projected 3D mask \( m_a \in \mathcal{M}_a\) should be assigned to the semantic label of the 2D mask \( m_b \in \mathcal{M}_b\) with the highest correspondence. 
An intuitive approach to associating the sets \(\mathcal{M}_a\) and \(\mathcal{M}_b\) is to apply a weighted bipartite matching algorithm. Given one mask from each bipartite set $m_a \in \mathcal{M}_a, m_b \in \mathcal{M}_b$, their weight is given by the inverse of the Jaccard Index \cite{fletcher2018jaccard}:
$$w(m_a, m_b) = \sum_i^h \sum_j^w \frac{|m_{a,ij} \cup m_{b,ij}|}{|m_{a,ij} \cap m_{b,ij}|}$$
However, we observe that SAM primarily proposes masks for instances rather than semantic classes. 
This means the 3D masks in \(\mathcal{M}_a\) often represent multiple instances or parts of the same class in \(\mathcal{M}_b\). This difference in the nature of masks introduces a mismatch in the cardinality of the sets \(\mathcal{M}_a\) and \(\mathcal{M}_b\), as there are generally several instances of each semantic class. Since bipartite matching can only efficiently assign each mask once, this mismatch complicates the process.
Switching to a generalized assignment problem (GAP) would allow multiple assignments but is known to be NP-hard \cite{cattrysse1992generalizedassignment}, therefore posing significant computational challenges. In contrast, bipartite matching can be efficiently solved using the Hungarian Algorithm \cite{kuhn1955hungarian}, which has cubic time complexity. Therefore, we opted not to switch to a GAP to maintain computational efficiency. Instead, we replicated the vertices in \(\mathcal{M}_b\) corresponding to the number of instances per class to match the instance-level correspondence required. This approach is solvable by the Jonker-Volgenant variant of the Hungarian Algorithm \cite{Jonker1987ASA, crouse2016implementing}, a version for non-square cost-matrices, ensuring a fast and effective assignment of semantic labels to our 3D-based class-agnostic masks.

A key advantage of our framework is its modularity: both the open-vocabulary 2D segmentation model and the 3D representation can be swapped without retraining. This flexibility stems from our use of class-agnostic segmentation masks, which decouple the 2D and 3D components. For instance, we validate this interchangeability by testing OpenSeg and OVSeg as class-aware 2D segmentation backbones (Tab. \ref{tab:ablation_study}), and the 3D representation could similarly be replaced to enhance class-agnostic mask accuracy. Furthermore, alternative mask assignment strategies—such as OpenMask3D’s feature-based mask classification \cite{takmaz2023openmask3d}—could be integrated in place of our bipartite matching mechanism. However, we prioritize computational efficiency and memory constraints, leading us to retain the lightweight bipartite assignment. Critically, no component changes necessitate retraining, making our approach adaptable to evolving segmentation architectures.

\begin{figure*}[h]
\begin{center}
\begin{tabular}{c c c}
    \includegraphics[width=0.31\textwidth]{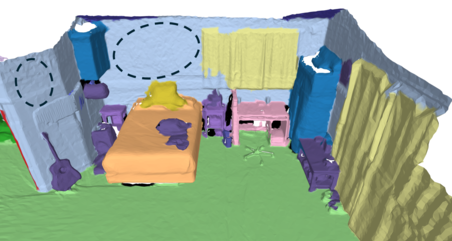} & 
    \includegraphics[width=0.3\textwidth]{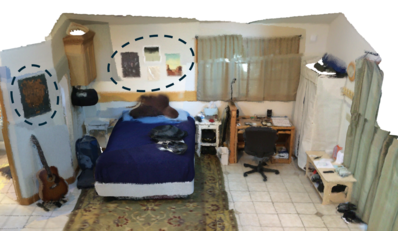} & 
    \includegraphics[width=0.29\textwidth]{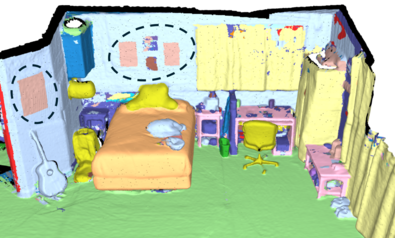} \\
    ScanNet Ground Truth & Colored Reconstruction & Segmentation (Ours)
\end{tabular}
\end{center}
\caption{\textbf{Shortcomings of the ScanNet GT}. Our Method accurately recognizes and segments the posters on the wall, but they are not represented in the provided ScanNet Ground Truth, therefore hurting our performance despite a more accurate segmentation of the scene.}
\label{fig:mask_images}
\end{figure*}

\subsection{Implementation Details}
Our method is implemented in Pytorch and runs on a single Nvidia RTX A5000 GPU with 24GB of memory. Due to the decoupled nature of our method and depending on the available setup and resources, multiple steps (e.g. training of the 3D Gaussian Representation and generation of the 2D segmentation masks) can easily be executed in parallel. The best-performing 2D segmentation model is OVSeg's biggest available model (Swin-Base + CLIP-ViT-L/14), which we utilize for inference only. Regarding the 3D Gaussian Spatting Reconstruction, we closely follow SAGA's approach with slight modifications to the clustering and scale parameters.
\section{Experiments}
\subsection{Datasets}
We evaluate our method both on synthetic data with the Replica Dataset \cite{straub2019replica} as well as real-world data with the ScanNet Dataset \cite{dai2017scannet}. Replica consists of high-quality scenes with realistic textures. It is well-suited for 3D open vocabulary semantic segmentation since it entails a long-tail distribution of small objects and very accurate semantic labels. We evaluate on the commonly used 8 scenes (\textit{office0, office1, office2, office3, office4, room0, room1, room2}). To ensure comparability to the baseline methods, we only evaluate on a subset of 200 of the original posed RGB-D images.
The annotations consist of 51 distinct class labels, and we follow OpenNeRF and split them further into (\textit{head, common, tail}) subsets, each consisting of 17 classes.
ScanNet consists of high-quality scans of indoor spaces, including significantly larger scenes than Replica. For evaluation, we use the 20-class subset of the NYUv2 40-label set since this is the setting in which the ground truth is given.
Note that our method does not use any of the provided ground truth semantic labels for training and is is not bound to the evaluation classes but able to segment any object or concept.

\begin{table*}
\centering
\resizebox{0.9\textwidth}{!}{%
\begin{tabular}{lcccccccc}
\toprule
 &
  \multicolumn{2}{c}{\textit{Total}} &
  \multicolumn{2}{c}{\textit{Head}} &
  \multicolumn{2}{c}{\textit{Common}} &
  \multicolumn{2}{c}{\textit{Tail}} \\ \cmidrule(r){2-3} \cmidrule(r){4-5} \cmidrule(r){6-7} \cmidrule(r){8-9}
 &
  \cellcolor[HTML]{EFEFEF}mIoU &
  mAcc &
  \cellcolor[HTML]{EFEFEF}mIoU &
  mAcc &
  \cellcolor[HTML]{EFEFEF}mIoU &
  mAcc &
  \cellcolor[HTML]{EFEFEF}mIoU &
  mAcc \\ \midrule
OpenSeg + Matching &
  \cellcolor[HTML]{EFEFEF}16.17 & 29.61 &
  \cellcolor[HTML]{EFEFEF}31.89 & 43.76 &
  \cellcolor[HTML]{EFEFEF}14.63 & 32.50 &
  \cellcolor[HTML]{EFEFEF}2.93 & 14.28 \\
OVSeg + Matching &
  \cellcolor[HTML]{EFEFEF}17.96 & 32.41 &
  \cellcolor[HTML]{EFEFEF}35.35 & 43.41 &
  \cellcolor[HTML]{EFEFEF}15.48 & 31.61 &
  \cellcolor[HTML]{EFEFEF}4.11 & \textbf{24.10} \\
OpenSeg + Assignment &
  \cellcolor[HTML]{EFEFEF}17.10 & 27.96 &
  \cellcolor[HTML]{EFEFEF}29.87 & 42.86 &
  \cellcolor[HTML]{EFEFEF}\textbf{18.95} & 33.61 &
  \cellcolor[HTML]{EFEFEF}3.47 & 9.05 \\ 
OVSeg + Assignment &
  \cellcolor[HTML]{EFEFEF}\textbf{19.91} & \textbf{33.11} &
  \cellcolor[HTML]{EFEFEF}\textbf{38.08} & \textbf{47.61} &
  \cellcolor[HTML]{EFEFEF}16.14 & \textbf{34.37} &
  \cellcolor[HTML]{EFEFEF}\textbf{6.69} & 19.31 \\ \bottomrule
\end{tabular}%
}
\caption{\textbf{Ablation Study on Replica.} Effect of different segmentation models and bipartite matching vs assignment (see Sec. \ref{sec:ablation_study})}
\label{tab:ablation_study}
\end{table*}

\begin{figure*}[ht]
\begin{center}
\begin{tabular}{c c c}
    \includegraphics[width=0.325\textwidth]{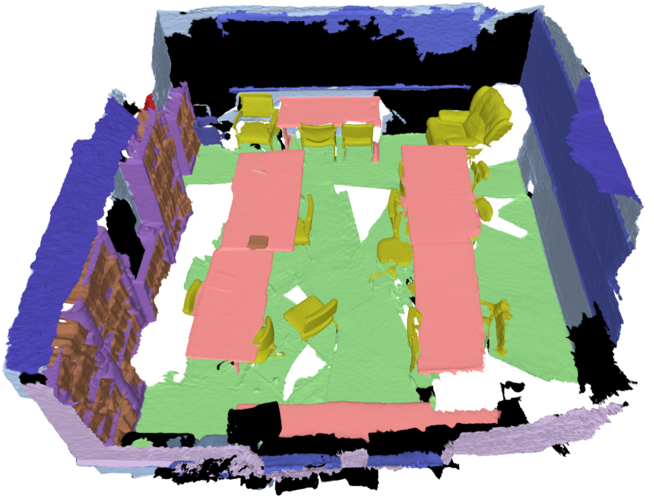} & 
    \includegraphics[width=0.3\textwidth]{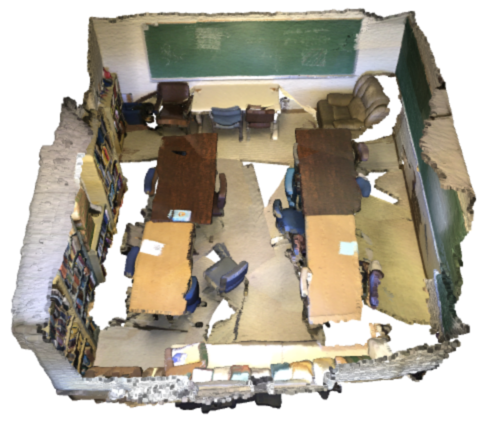} & 
    \includegraphics[width=0.3\textwidth]{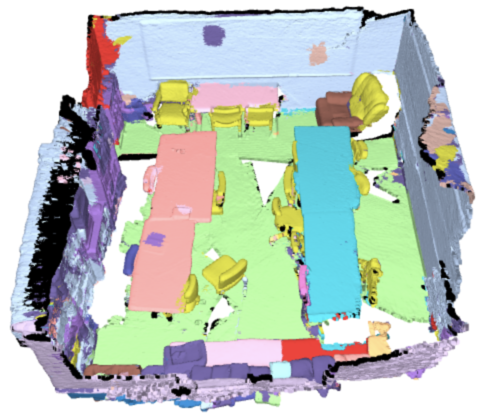} \\
    ScanNet Ground Truth & Colored Reconstruction & Segmentation (Ours) \\
    \includegraphics[width=0.3\textwidth]{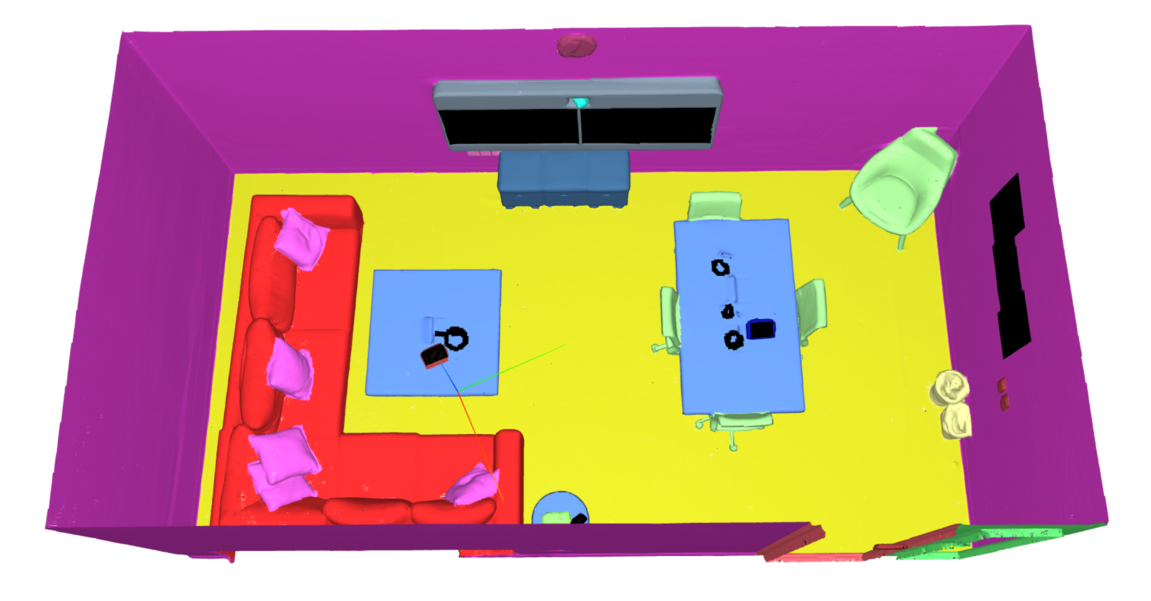} & 
    \includegraphics[width=0.3\textwidth]{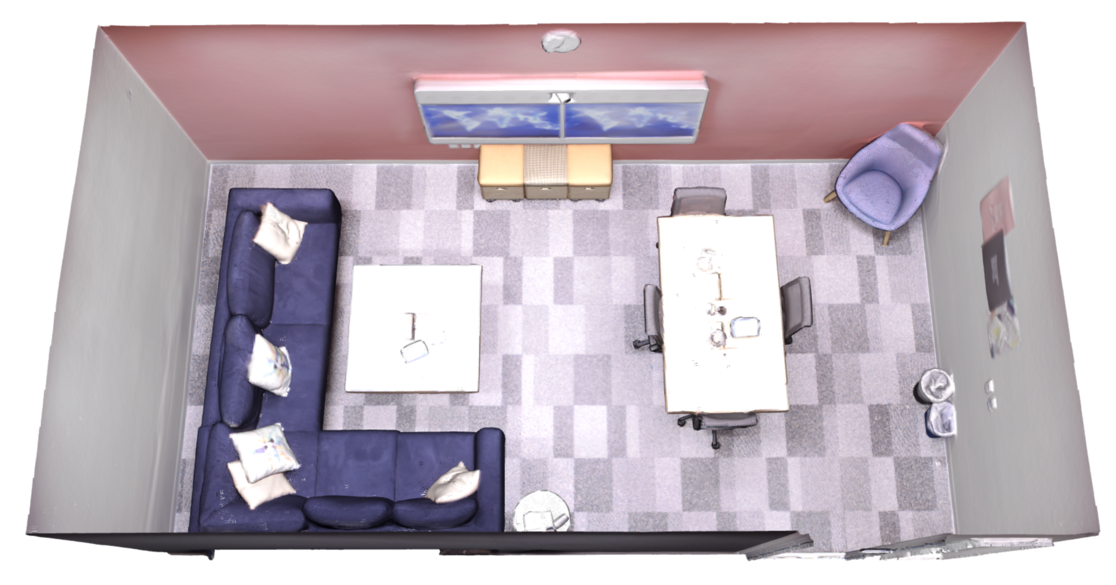} & 
    \includegraphics[width=0.3\textwidth]{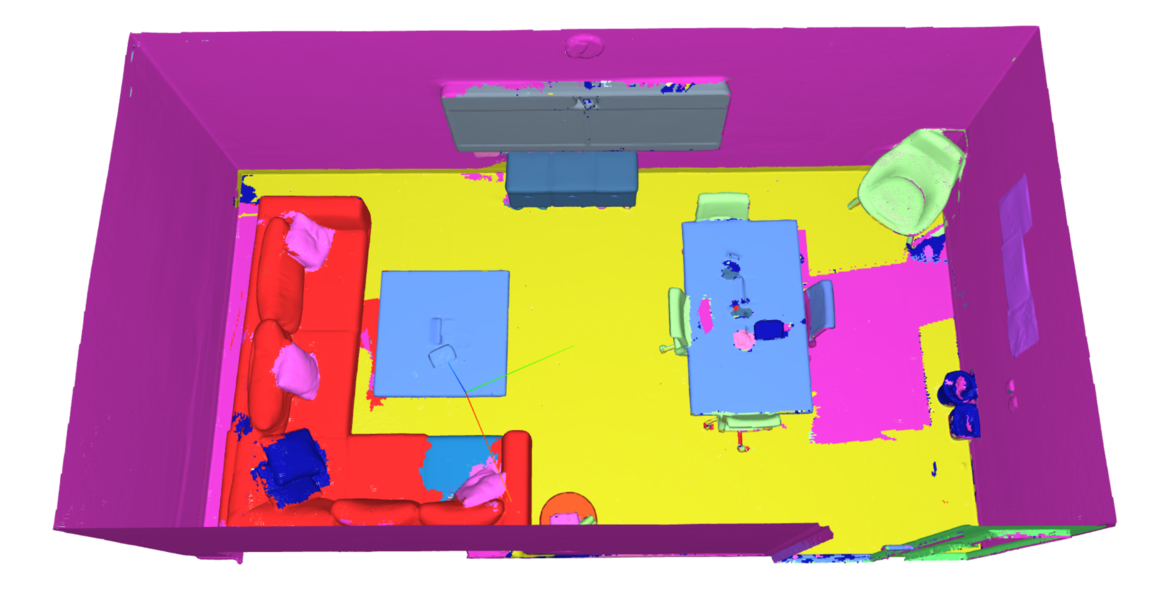} \\
    Replica Ground Truth & Colored Reconstruction & Segmentation (Ours)
\end{tabular}
\end{center}
\caption{\textbf{Further Results on ScanNet (\textit{scene0030\_01}) and Replica (\textit{office2})}}
\label{fig:further_results}
\end{figure*}

\subsection{Metrics}
For a quantitative evaluation of our method, we project our semantic predictions back to the given annotated point clouds and follow OpenNeRF and ScanNet to report the \textit{mean intersection over union (mIoU)} and \textit{mean accuracy (mAcc)} for the whole scene as well as the subsets. 

\subsection{Synthetic Data: Replica Dataset} 
When comparing our results to pipelines based only on 2D class-aware segmentation features (e.g. OpenNeRF), we see that our masks are more accurate. This happens, in particular, if the scene has some shadows. This improvement can likely be accounted for by the additional availability of 3D geometry, making classification easier.
Compared to OpenNeRF, we can observe that our method achieves less scattered results in large areas. These artifacts are part of the MLP and the rendering function which is based on ray-tracing. In contrast, the explicit geometry in Gaussian Splatting as an underlying representation ensures consistency for these areas. Borders of smaller objects, such as pillows and blankets, are sharper compared to OpenNeRF.
Note that decoupling the 3D segmentation proposals from the class-aware segmentation masks allows us to simultaneously perform instance segmentation. Each pillow was assigned "pillow" as a label, but the clusters were identified separately before the assignment (see Fig. \ref{fig:results_img}).
Our method outperforms the NeRF-based baseline, OpenNeRF, in all but one subfield (\textit{common mIoU}) despite using a completely different architecture that significantly increases the modularity (see Tab. \ref{tab:segmentation_results}). The effect of differing open-vocabulary segmentation models is apparent when comparing OpenSeg to OVSeg, which offers a notable difference in tail-class performance (see Tab. \ref{tab:ablation_study}). This means the segmentation performance is still heavily influenced by the underlying 2D Segmentation Foundation Model, further reinforcing our approach of decoupling the 3D segmentation pipeline to ensure modularity and adaptability to this fast-evolving field. 

\subsection{Real-World Data: ScanNet v2}
OpenNeRF does not report any quantitative measures on real-world data. To validate our performance from synthetic data on real-world data, we evaluate both our method and OpenNeRF on four scenes from ScanNet v2, the commonly used \textit{scene0000\_00} from the category \textit{Apartment} as well as one randomly picked scene from \textit{Classroom} (\textit{scene0030\_01}) and two from \textit{Bathroom} (\textit{scene0062\_00} and \textit{scene0100\_01}). It is important to note that these scenes initially contain 5578, 1648, 730, and 1120 posed RGB-D images. To challenge the effectiveness of our method and compare it to synthetic data, we only utilized 200 images for reconstruction and segmentation, meaning only a fraction of the available data for each scene. Additionally, we also don't utilize any of the available ScanNet annotations for training but rather perform our segmentation in a zero-shot manner. Results can be seen in Fig. \ref{fig:mask_images}, \ref{fig:further_results} and Tab. \ref{tab:scannet_results}.

\begin{table}[H]
\centering
\begin{tabular}{lcc}
\toprule
 & \cellcolor[HTML]{EFEFEF} mIoU & mAcc \\
\midrule
OpenNeRF \cite{engelmann2024opennerf} & \cellcolor[HTML]{EFEFEF} 49.5 & 62.7\\
Ours & \cellcolor[HTML]{EFEFEF} \textbf{55.1} & \textbf{63.5} \\
\midrule
\textcolor{mygray}{MinkowskiNet \cite{choy20194d}} & \cellcolor[HTML]{EFEFEF} \textcolor{mygray}{69.0} & \textcolor{mygray}{77.5} \\
\textcolor{mygray}{OpenScene (LSeg) \cite{peng2023openscene}} & \cellcolor[HTML]{EFEFEF} \textcolor{mygray}{54.2} & \textcolor{mygray}{66.6} \\
\textcolor{mygray}{OpenScene (OpenSeg) \cite{peng2023openscene}} & \cellcolor[HTML]{EFEFEF} \textcolor{mygray}{47.5} & \textcolor{mygray}{70.7} \\
\bottomrule
\end{tabular}
\caption{\textbf{Semantic segmentation results on 
the choosen scenes of ScanNet v2 utilizing 200 images, a fraction of the original amount.} The grey values provide a reference; MinkowskiNet is one of the strongest fully-supervised approaches. OpenScene is zero-shot and utilizes point clouds, i.e. sparse geometry.}
\label{tab:scannet_results}
\end{table}

We observe that there are some areas where segmentation masks are accurate but the assignment of the correct label is unsuccessful. This indicates that our mask proposal is successful, but the underlying 2D foundation model may be unable to assess a given object accurately. A strength that we can observe is that even with limited data, our method is able to pick up long-tail classes that are not even represented in the ground-truth annotations, as seen with the posters on the wall (see Fig. \ref{fig:mask_images}). Keeping the amount of data equal, we continue outperforming OpenNeRF on ScanNet. While our performance is very competitive with respect to mIoU, we are slightly inferior in terms of mAcc with respect to MinkowskiNet and OpenScene. One likely cause of this observation could be that we only utilize a subset of the given images to evaluate a scene.

\subsection{Ablation Study}
\label{sec:ablation_study}

\paragraph{Effect of different segmentation models.}
Due to our modular architecture, we can easily swap between different 2D Foundation Segmentation Models. We conducted a small ablation study utilizing both OpenSeg and OVSeg (see Tab. \ref{tab:ablation_study}).
A notable difference in tail-class performance is apparent. Furthermore, our pipeline can be adapted to different tasks by switching the underlying 2D segmentation model to fit the user's specific needs.

\paragraph{Bipartite Matching vs. Bipartite Assignment.}
As mentioned previously, a bipartite matching formulation faces the challenge that every class can only be assigned once. We hypothesized that relaxing the bipartite matching formulation by introducing duplicates of semantic masks would mitigate the risk of misalignment between instances in the 3D clustering and classes in our 2D foundation model. To test this hypothesis, we have tested both 2D foundation models with a bipartite matching and the relaxed assignment formulation. Our ablation study (as shown in Tab. \ref{tab:ablation_study}) confirms our hypothesis and indicates that OVSeg, in combination with bipartite assignment, is the most promising approach.
\section{Discussion}

\subsection{Instance and Part Segmentation.}
\begin{figure}[h]
    \centering
    \begin{minipage}{0.48\columnwidth}
        \centering
        \includegraphics[width=\linewidth]{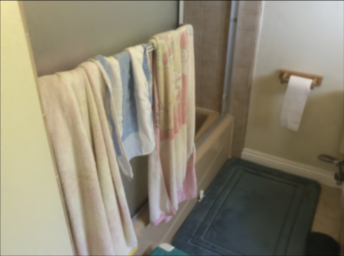}
    \end{minipage}
    \hfill
    \begin{minipage}{0.48\columnwidth}
        \centering
        \includegraphics[width=\linewidth]{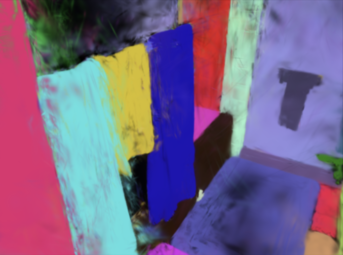}
    \end{minipage}
    
    \vspace{0.3cm} 
    
    \begin{minipage}{0.48\columnwidth}
        \centering
        \includegraphics[width=\linewidth]{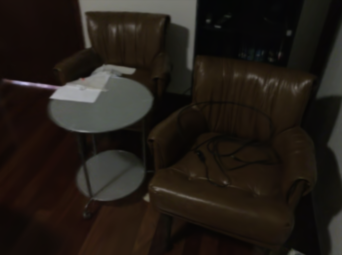} 
    \end{minipage}
    \hfill
    \begin{minipage}{0.48\columnwidth}
        \centering
        \includegraphics[width=\linewidth]{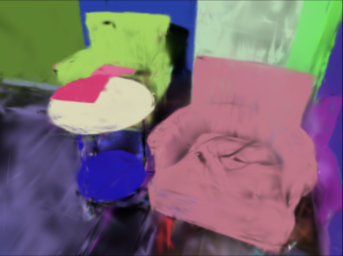} 
    \end{minipage}
    
    \caption{\textbf{Visualization of class-agnostic masks.} Our mask proposal tends to propose instances, as demonstrated by the three separately identified towels and two armchair instances.}
    \label{fig:instance_examples_masks}
\end{figure}

As mentioned previously and demonstrated in Figure \ref{fig:instance_examples_masks}, our approach predominantly learns masks for instances, enabling precise instance segmentation capabilities. 

\begin{figure}[h]
    \centering
    \begin{minipage}{0.225\columnwidth}
        \centering
        \includegraphics[width=\linewidth]{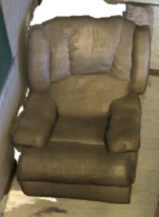}
    \end{minipage}
    \hfill
    \begin{minipage}{0.235\columnwidth}
        \centering
        \includegraphics[width=\linewidth]{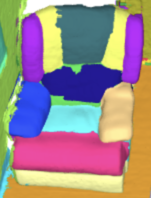}
    \end{minipage}
    \begin{minipage}{0.243\columnwidth}
        \centering
        \includegraphics[width=\linewidth]{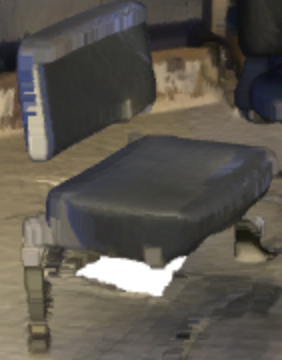}
    \end{minipage}
    \hfill
    \begin{minipage}{0.255\columnwidth}
        \centering
        \includegraphics[width=\linewidth]{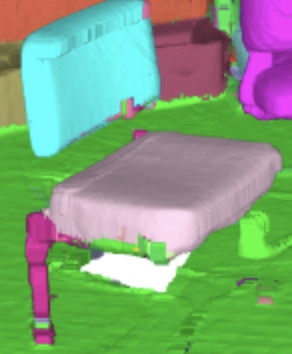}
    \end{minipage}
    \caption{\textbf{Part Segmentation.} For objects with clearly separable parts, our approach tends to propose masks that correspond to part segmentation.}
    \label{fig:part_seg}
\end{figure}

Moreover, adjusting the scale parameter $s$ based on the estimated number of visible objects in a scene is a significant advantage. This adaptability allows for the consideration of smaller objects without necessitating retraining of the model, thanks to scale-conditioned affinity features. The modular nature of our approach further enhances its utility, allowing for substituting the mask proposal network with one better suited for more fine-grained tasks, such as part segmentation.
The underlying SAM masks that we use for our architectures are not geared toward a specific goal like instance or part segmentation but still demonstrate the ability to perform both tasks as demonstrated in Figure \ref{fig:part_seg}. Even though our approach is not specialized to perform instance or part segmentation and we are mainly interested in the correct aggregation to class level, this additional capability can provide useful supplementary information.  

In conclusion, our methodology offers distinct advantages over our baselines by enabling instance or even part segmentation without needing network retraining or architectural redesign, thus providing a flexible and robust solution for diverse segmentation tasks.

\subsection{Limitations} 
\paragraph{Bipartite Assignment is not optimal.} Matching multiple projected instance proposals to 2D segmentation masks ideally requires a generalized assignment instead of a bipartite one. To account for this weakness, future work could replace the matching step and directly perform the classification step on the mask proposals. While such approaches exist in 2D \cite{cheng2022mask2former}, they must be trained on ground-truth data, requiring significant computational resources for training and large datasets to be generalizable across domains. In 3D, the high computational requirements of such approaches \cite{takmaz2023openmask3d, yin2024sai3d} are a concern that needs to be addressed. Another approach that is left for future work and is in line with our architecture involves designing a more flexible assignment algorithm that adjusts to scene size and object counts both in general and in individual frames to further increase the robustness of the assignment formulation. 

\paragraph{Tail-class performance is limited by the 2D foundation model.} The above approach also addresses the reliance on an almost perfect match between our masks and the compared 2D masks. Classes that the 2D model fails to recognize but are identified by the 3D clustering cannot be accurately labeled. Thus, one of our key advantages—accurately identifying long-tail classes using a combination of 3D geometry and segmentation features—is compromised if the class-aware foundation model underperforms. 

\begin{figure}[h]
    \centering
    \includegraphics[width=0.75\linewidth]{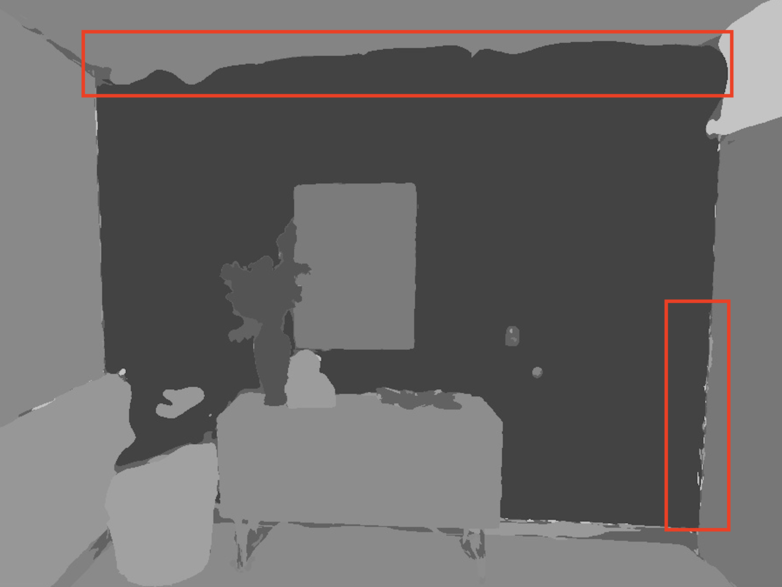}
    \caption{\textbf{Limitations.} The nature of 3D Gaussians prevents clear edges when segmenting (top). Additionally, we can observe low-connectivity clusters with few pixels between the two walls (right).}
    \label{fig:edge_failure}
\end{figure}

\paragraph{Gaussians vs. Sharp Edges.} Gaussians, due to their inherent spherical nature, sometimes struggle to accurately segment objects with sharp edges. This limitation leads to imprecise boundaries and overlaps in the segmentation output as seen in Figure \ref{fig:edge_failure}. There are alternative approaches and modifications to Gaussian-based models to better handle complex geometries with sharp edges and mitigate this issue. For instance, Hu et al. \citep{hu2024sagd} refine Gaussian segmentation by decomposing Gaussians to address this shortcoming, enabling Gaussian-based segmentation methods for domains in which accuracy is crucial.

\paragraph{Outlier Clusters.} The proposed feature extraction represents a relatively novel method to leverage 3D Gaussians for segmentation mask proposal, offering a new perspective in the realm of 3D segmentation. However, an observed challenge (see Fig. \ref{fig:edge_failure}) with this approach is its sensitivity to outlier clusters that exhibit low connectivity. These clusters, which do not conform to the expected connectivity patterns, can adversely affect the overall accuracy of the segmentation. To counteract this, we implement a post-processing step that systematically identifies and eliminates clusters with low connectivity. Despite this measure, challenges persist when clusters that marginally exceed the connectivity threshold remain in the data. Such clusters continue to pose a problem, indicating the need for more sophisticated strategies to ensure robust segmentation.
\section{Conclusion}
We presented DCSEG, a decoupled pipeline for open-vocabulary 3D semantic segmentation that is simultaneously able to segment parts and instances that can be aggregated to classes without the need for retraining. We utilize 3D Gaussian Splatting as an underlying scene representation. This alternative to NeRF-based approaches shows improved results while being computationally more efficient. Additionally, we provide a way to approximate a general assignment by matching clusters over multiple image pairs and propose a modular framework that can easily be adapted if novel methods for either 3D instance proposals or 2D open-vocabulary segmentation become available.
{
    \small
    \bibliographystyle{ieeenat_fullname}
    \bibliography{main}
}

\appendix

\end{document}